\documentclass[
]{ceurart}
\usepackage{algorithmic}
\usepackage{xspace}
\usepackage{csquotes}
\begin{document}

\copyrightyear{2020}
\copyrightclause{Copyright for this paper by its authors.\\
  Use permitted under Creative Commons License Attribution 4.0
  International (CC BY 4.0).}

\conference{FIRE 2020: Forum for Information Retrieval Evaluation,
  December 16--20, 2020, IDRBT, Hyderabad,India}

\title{CUSATNLP@HASOC-Dravidian-CodeMix-FIRE2020: Identifying Offensive Language from Manglish Tweets}

\author[1]{Sara Renjit}[%
orcid=0000-0002-9932-2039,
]
\ead{sararenjit.g@gmail.com}
\address[1]{Department of Computer Science, Cochin University of Science and Technology, Kerala, India}

\author[2]{Sumam Mary Idicula}
\address[2]{Department of Computer Science, Cochin University of Science and Technology, Kerala, India}
\ead{sumam@cusat.ac.in}

\begin{abstract}
  With the popularity of social media, communications through
blogs, Facebook, Twitter, and other platforms have increased. Initially, English was the only medium of communication. Fortunately, now we can communicate in any language. It has led to people using English and their own native or mother tongue language in a mixed form. Sometimes, comments in other languages have English transliterated format or other cases; people use the intended language scripts. Identifying sentiments and offensive content from such code mixed tweets is a necessary task in these times. We
present a working model submitted for Task2 of the sub-track HASOC Offensive Language Identification- DravidianCodeMix in Forum for Information Retrieval Evaluation, 2020. It is a message level classification task. An embedding model-based classifier identifies offensive and not offensive comments in our approach. We applied this method in the Manglish dataset provided along with the sub-track.
\end{abstract}

\begin{keywords}
   Offensive Language \sep
  Social Media Texts \sep
  Code-Mixed \sep
  Embeddings \sep
  Manglish 
\end{keywords}

\maketitle

\section{Introduction}
As code-mixing has become very common in present communication media, detecting offensive content from code mixed tweets and comments is a crucial task these days \cite{9074379,9074205}. Systems developed to identify sentiments from the monolingual text are not always suitable in a multilingual context. Hence we require efficient methods to classify offensive and non-offensive content from multilingual texts. In this context, two tasks are part of the HASOC FIRE 2020 sub-track \cite{hasoc}. The first task deals with the message-level classification of code mixed YouTube comments in Malayalam, and the second task deals with the message-level classification of tweets or Youtube comments in Tanglish and Manglish (Tamil and Malayalam using written using Roman characters). Tamil and Malayalam languages are Dravidian languages spoken in South India \cite{chakravarthi2018improving,chakravarthi2019wordnet,chakravarthi2019comparison,raja2020survey}.

The following sections explain the rest of the contents: Section 2 presents related works in offensive content identification. Task description and dataset details are included in Section 3. Section 4 explains the methodology used. Section 5 relates
to experimental details and evaluation results. Finally, Section 6 concludes the work.

\section{Related Works}
We discuss works done related to offensive content identification in the past few years. Offensive content detection from tweets is part of some conferences as challenging tasks. In 2019, SemEval (Semantic Evaluation) \cite{zampieri2019semeval} conducted three tasks, out of which the first task was the identification of offensive and non-offensive comments in English tweets. The dataset used was OLID. It has 14000 tweets annotated using a hierarchical annotation model. The training set has 13240 tweets, and a test set has 860 tweets. They used different methods like Convolutional Neural Networks (CNN), Long Short Term Memory (LSTM), LSTM with attention, Embeddings from Language Models (ELMo), and Bidirectional Encoder Representations from Transformers (BERT) based systems. Also, few teams attempt traditional machine learning approaches like logistic regression and support vector machine (SVM) \cite{zampieri2019semeval}.
A new corpus developed for sentiment analysis of code-mixed text in Malayalam-English is detailed in \cite{chakravarthi-etal-2020-sentiment}.
SemEval in 2020 presented offensive language identification in multilingual languages named as OffensEval 2020, as a task in five languages, namely English, Arabic, Danish, Greek, and Turkish \cite{zampieri2020semeval}. The OLID dataset mentioned above extends with more data in English and other languages. Pretrained embeddings from BERT-transformers, ELMo, Glove, and Word2vec are used with models like BERT and its variants, CNN, RNN, and SVM.

The same task was conducted for Indo-European languages in FIRE 2019 for English, Hindi, and German. The dataset was created by collecting samples from Twitter and Facebook for all the three languages. Different models such as LSTM with attention, CNN with Word2vec embedding, BERT were used for this task. In some cases, top performance resulted from traditional machine learning models, rather than deep learning methods for languages other than English \cite{hasoc}. Automatic approaches for hate speech also includes keyword-based approaches and TF-IDF based multiview SVM, which is described in \cite{macavaney2019hate}.

\section{Task Description \& Dataset}

HASOC sub-track, Offensive Language Identification of Dravidian CodeMix \cite{hasocdravidian--acm} \cite{dravidiansentiment-acm} \cite{dravidiansentiment-ceur} consists of two tasks: message level classification of YouTube comments in code mixed Malayalam (Task1) and message level classification of Twitter comments in Tamil and Malayalam written using Roman characters(Task2). This task is in continuation with the HASOC, 2019 task as in \cite{mandl2019overview}.
The details about corpus creation and its validation are detailed in \cite{chakravarthi-etal-2020-corpus}. Orthographic information is also utilized while creating the corpus, as mentioned in  \cite{chakravarthi2020leveraging}. This paper discusses the methods used for Task 2 classification of twitter comments in Manglish text. The training dataset consists of comments in two different classes for the task, as shown in Table~\ref{datatab}.
\begin{table}
    \caption{Dataset statistics for Task2: Malayalam Code-Mix}\label{datatab}
	\centering	
	\begin{tabular}{|l|c|}
		\hline
		\textbf{Class} &  \textbf{Count} \\
		\hline
		Offensive & 1953 \\
		\hline
		Not Offensive & 2047\\
		\hline
		Total & 4000\\
		\hline
	\end{tabular}
\end{table}

\section{Proposed Methods}
\begin{figure}
	\includegraphics[width=\textwidth]{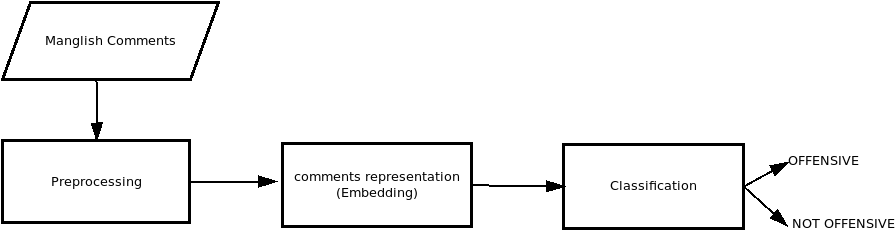}
	\caption{Design of the proposed methods} \label{sysdesign}
\end{figure}
We present three submissions experimenting with three different methods based on the general system design, as in Figure~\ref{sysdesign}. The offensive language identification system consists of the following stages:
\begin{enumerate}
	\item Preprocessing: This stage includes preprocessing texts based on removing English stopwords present in the Manglish comment texts, URLs defined with prefix \textquote{www} or \textquote{https}, usernames with the prefix \textquote{@}, hash in hashtags, repeated characters, unwanted numbers. All text is converted to lowercase and tokenized into words. Tweet preprocessor \footnote{https://pypi.org/project/tweet-preprocessor} for cleaning tweets is used to remove hashtags, URLs, and emojis. 
	
	\item Comments representation: Here, we embed the comments based on two embedding mechanisms: 
		\begin{itemize}
			\item Using Keras embedding\footnote{https://keras.io/api/layers/core\_layers/embedding}, we represent the sentences using one-hot representation, adequately padded to a uniform length, and passed to Keras embedding layer to produce 50-dimensional sentence representations.
			
			\item Using paragraph vector, an unsupervised framework that learns distributed representations from texts of variable length \cite{le2014distributed}. The paragraph vector is an algorithm that uses  Word2vec based word vector representations \cite{mikolov2013efficient}.
			
		\end{itemize}
	\item Classification: In this step, the comments represented as n-dimensional vectors are trained with the following network parameters:
	\begin{itemize}
		\item System A: Classifier with an LSTM layer and recurrent dropout(0.2) followed by a dense layer with sigmoid activation and binary cross-entropy classifies comments as offensive or not offensive in 5 epochs.
		
		\item System B: Classifier with three dense layers with Relu activation and the final layer is dense with sigmoid activation and binary cross-entropy and trained for 50 epochs for classification.
	
        \item System C: Combination of two classifiers: We use a mathematical combination of predictions from both classifiers to produce the third submission results, based on a decision function. Prob(X) denotes the probability values output by system X, and Pred(X) denotes the predicted class of System X.
    
        \subsubitem \textbf{Decision function:}
    	    \begin{algorithmic}
    		    \IF{Pred(A) == Pred(B)}
    		    \STATE Pred(C)= Pred(A or B)
    		    \ELSIF{Prob(A)+Prob(B) greater than 1}
    		    \STATE Pred(C) = \textquote{Offensive}
    		    \ELSE
    		    \STATE Pred(C) = \textquote{Not Offensive}
    		    \ENDIF
    	    \end{algorithmic}
   
    \end{itemize}
\end{enumerate}
\section{Experimental Results}
The proposed system\footnote{https://github.com/SaraRenG/Code1} is trained on 4000 comments from the training set and tested on  1000 comments. The weighted average F1-score is used for evaluation as it is an imbalanced binary classification task. Table~\ref{sysA} shows the performance of System A based on Keras embedding and LSTM layer, Table~\ref{sysB} shows System B's results based on document embedding using Doc2Vec, and Table~\ref{sysC} is the result of the combined classifier based on mathematical logic. We use precision, recall, and F1-score as evaluation metrics. The weighted average F1-score is more significant as it handles class imbalance.  Its presence in the data sample weights each class score, showing a balance between precision and recall. We calculate these metrics using the Scikit-Learn\footnote{https://scikit-learn.org/stable/modules/generated/sklearn.metrics.classification\_report.html} package. All the results show an average performance, which shows the scope for improvement.
\begin{table}
    \caption{Evaluation results for System A: based on Keras Embedding}\label{sysA}
	\centering	
	\begin{tabular}{| c | c | c | c | c |}
		\hline
		\textbf{System A} &\textbf{ Precision} & \textbf{ Recall} & \textbf{F1-score} &\textbf{ Support} \\
		\hline
		NOT  &     0.52 &     0.60 &     0.56 &      488\\
		OFF   &    0.55  &    0.48  &    0.51  &    512\\
		\hline
		micro avg  &  0.54  &   0.54  &   0.54  &    1000\\
		macro avg  &  0.54  &   0.54  &   0.53  &    1000\\
		weighted avg  &  0.54  &  0.54 &  0.53  &  1000\\
		\hline
	\end{tabular}
\end{table}
\begin{table}
    \caption{Evaluation results for System B: based on Doc2Vec}\label{sysB}
	\centering
	\begin{tabular}{|c|c|c|c|c|}
		\hline
		\textbf{System B}& \textbf{Precision} &\textbf{  Recall} &\textbf{ F1-score} &\textbf{ Support} \\
		\hline
		NOT  &    0.49 &  0.67   &   0.56  &   488\\
		OFF  &    0.51  &    0.32  &    0.40 &   512 \\
		\hline
		micro avg  &  0.49  & 0.49 &  0.49 &   1000 \\
		macro avg  &  0.50  & 0.50 &  0.48 &    1000\\
		weighted avg &   0.50  &  0.49  &  0.48   & 1000\\
		\hline
	\end{tabular}
\end{table}

\begin{table}
    \caption{Evaluation results for System C: based on Decision function}\label{sysC}
	\centering	
	\begin{tabular}{|c|c|c|c|c|}
		\hline
		\textbf{System C}& \textbf{Precision} & \textbf{ Recall} & \textbf{F1-score} &\textbf{ Support} \\
		\hline
		NOT &    0.53   &  0.49 &   0.51 &   488 \\
		OFF  &   0.55   &   0.58  &   0.57  &   512\\
		\hline
		micro avg &   0.54 &  0.54 &   0.54  &  1000\\
		macro avg   &  0.54 &   0.54 &  0.54  &  1000\\
		weighted avg  &  0.54 &  0.54  &  0.54  &  1000\\
		\hline
	\end{tabular}
\end{table}
\section{Conclusion}
This paper presents offensive content identification systems for Manglish tweets or comments. It is as part of the HASOC sub-track in FIRE, 2020. We implemented simple methods using sentence representations and binary classification using neural networks. Significant challenges in this task are the representation of text in Manglish(Malayalam written using Roman characters), it's embedding without losing much information, which we can further improve in future attempts.


\bibliography{WorkingNote}

\end{document}